\documentclass[11pt,a4paper]{article}
\usepackage[hyperref]{acl2020}
\usepackage{times}
\usepackage{latexsym}

\usepackage{microtype}
\usepackage{amsmath}
\newcommand{\xhdr}[1]{{\noindent\bfseries #1}.}

\newcommand{\cut}[1]{}

\usepackage{graphicx}
\usepackage{multirow}
\usepackage[caption=false]{subfig}
\usepackage{todonotes}
\usepackage{hyperref}
\usepackage{enumitem}

\newcommand{\matr}[1]{\mathbf{#1}}
\newcommand{\vect}[1]{\mathbf{#1}}

\newcommand{\vc}[0]{\vect{c}}

\newcommand{\vh}[0]{\vect{h}}

\newcommand{\mW}[0]{\matr{W}}
\newcommand{\mD}[0]{\matr{D}}

\newcommand\boldblue[1]{\textcolor{blue}{\textbf{#1}}}

\usepackage{amsmath}
\usepackage{tikz}
\usepackage{mathdots}
\usepackage{yhmath}
\usepackage{cancel}
\usepackage{color}
\usepackage{siunitx}
\usepackage{array}
\usepackage{multirow}
\usepackage{amssymb}
\usepackage{gensymb}
\usepackage{tabularx}
\usepackage{booktabs}
\usetikzlibrary{fadings}
\usetikzlibrary{patterns}
\usetikzlibrary{shadows.blur}

\newcommand{\dtm}{{\sc{MaUde}}}
\newcommand{\dtma}{{\textbf{M}etric for \textbf{a}utomatic \textbf{U}nreferenced \textbf{d}ialogue \textbf{e}valuation}}

\aclfinalcopy 
\def\aclpaperid{753} 

\title{Instructions for ACL 2020 Proceedings}

\date{}
\begin{document}
\title{Learning an Unreferenced Metric for Online Dialogue Evaluation}

\author{Koustuv Sinha\thanks{\ \  Corresponding author: koustuv.sinha@mail.mcgill.ca. Code for reproducing the experiments are available at \href{https://github.com/facebookresearch/online_dialog_eval}{https://github.com/facebookresearch/online\_dialog\_eval}.}, \textsuperscript{1,2,3}
  Prasanna Parthasarathi, \textsuperscript{1,2}
  Jasmine Wang, \textsuperscript{1} \\ 
  {\bf Ryan Lowe, \textsuperscript{1,2,4}}
  {\bf William L. Hamilton, \textsuperscript{1,2}}  and
  {\bf Joelle Pineau \textsuperscript{1,2,3}} \\
  \textsuperscript{1} School of Computer Science, McGill University, Canada \\
  \textsuperscript{2} Quebec Artificial Intelligence Institute (Mila), Canada \\
  \textsuperscript{3} Facebook AI Research (FAIR), Montreal, Canada\\
  \textsuperscript{4} OpenAI
  \\
}


\maketitle
\begin{abstract}
Evaluating the quality of a dialogue interaction between two agents is a difficult task, especially in open-domain chit-chat style dialogue.  There have been recent efforts to develop automatic dialogue evaluation metrics, but most of them do not generalize to unseen datasets and/or need a human-generated reference response during inference, making it infeasible for online evaluation. Here, we propose an {\em unreferenced} automated evaluation metric that uses large pre-trained language models to extract latent representations of utterances, and leverages the temporal transitions that exist between them. We show that our model achieves higher correlation with human annotations in an online setting, while not requiring true responses for comparison during inference.
\end{abstract}


\section{Introduction}
Recent approaches in deep neural language generation have opened new possibilities in dialogue generation \cite{serban2017hierarchical, weston2018retrieve}. Most of the current language generation efforts are centered around language modelling or machine translation \cite{ott2018scaling}, which are evaluated by comparing directly against the reference sentences. In dialogue, however, comparing with a single reference response is difficult, as there can be many reasonable responses given a context that have nothing to do with each other \cite{liuHowNOTEvaluate2016}. Still, dialogue research papers tend to report scores based on \textit{word-overlap} metrics from the machine translation literature (e.g. BLEU \cite{papineni2002bleu}, METEOR \cite{denkowski2014meteor}).  However word-overlap metrics aggressively penalize the generated response based on lexical differences with the ground truth and correlate poorly to human judgements \cite{liuHowNOTEvaluate2016}. 

\begin{figure}
    \centering
    \resizebox{0.5\textwidth}{!}{
    \includegraphics[]{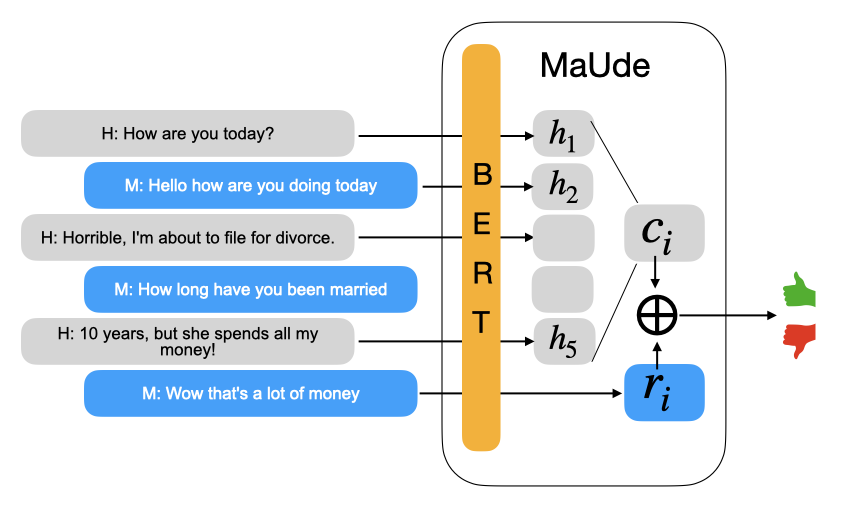}}
    \caption{
   Model architecture for MaUdE, which is an unsupervised unreferenced metric for dialog evaluation.}
    \label{fig:worlds}
\end{figure}



One can build dialogue evaluation metrics in two ways: \textit{referenced} metrics, which compare the generated response with a provided ground-truth response (such as the above word-overlap metrics), or an \textit{unreferenced} metrics, which evaluate the generated response without any such comparison. 
\citet{loweAutomaticTuringTest2017a} propose a \textit{learned referenced} metric named ADEM, which learns an alignment score between context and response to predict human score annotations. However, since the score is trained to mimic human judgements, it requires collecting large-scale human annotations on the dataset in question and cannot be easily applicable to new datasets \cite{loweRetrospectiveAutomaticTuring2019}.  

Recently, \citet{taoRUBERUnsupervisedMethod2017} proposed a hybrid referenced-unreferenced metric named RUBER, where the metric is trained without requiring human responses by bootstrapping negative samples directly from the dataset. 
However, referenced metrics (including RUBER, as it is part referenced) are not feasible for evaluation of dialogue models in an \textit{online} setting---when the model is pitched against a human agent (model-human) or a model agent (model-model)---due to lack of a reference response. In this setting, models are usually evaluated directly by humans, which is costly and requires careful annotator training \cite{li2019acute}. 

The contributions of this paper are (1) a completely unsupervised unreferenced metric \dtm{} (\dtma{}), which leverages state-of-the-art pre-trained language models \cite{devlinBERTPretrainingDeep2018, sanh2019distilbert}, combined with a novel discourse-structure aware text encoder and contrastive training approach; and (2)
 results showing that \dtm{} has good correlation with human judgements.

\section{Background}
We consider the problem of evaluating the response of a dialogue system, where an agent is provided with a sequence of sentences (or utterances) $c = \{u_1,u_2,...,u_n\}$ (termed as \textit{context}) to generate a \textit{response} $r=u_{n+1}$. Each utterance, $u_i$, can be represented as a set of words $u_i = \{w_1,w_2,...,w_n\}$. An utterance $u_i$ can be represented as a vector as $\vh_i = f_e(u_i)$, where $f_e$ is an encoder that encodes the words into a fixed vector representation. 

This work focuses on the evaluation of \textit{generative neural dialogue models}, which typically consist of an encoder-decoder style architecture that is trained to generate $u_{n+1}$ word-by-word \cite{serban2017hierarchical}.
The response of a generative model is typically evaluated by comparing with the ground-truth response using various automatic word-overlap metrics, such as BLEU or METEOR. These metrics, along with ADEM and RUBER, are essentially \textit{single-step} evaluation metrics, where a score is calculated for each context-response pair. If a dialogue $D_i$ contains $n$ utterances, we can extract $n-1$ context-response pairs : $(c_1  : \{u_1\}, r_1 : \{u_2\}), (c_2 : \{u_1, u_2\}, r_2 : \{u_3\}), \ldots, (c_{n-1} : \{u_1 \ldots u_{n-1}\}, r_{n-1} : u_n)$. In this paper, we are interested in devising a scalar metric that can evaluate the quality of a context-response pair: $\text{score}(c_i,r_i) = \mathcal{R} \in (0,1)$. 
A key benefit of this approach is that this metric can be used to evaluate online and also for better training and optimization, as it provides partial credit during response generation. 

\section{Proposed model}
\label{sec:model}




We propose a new model, \dtm{}, for online unreferenced dialogue evaluation. 
We first describe the general framework behind \dtm{}, which is inspired by the task of measuring alignment in natural language inference (NLI) \cite{williams2017broad}. It involves training text encoders via noise contrastive estimation (NCE) to distinguish between valid dialogue responses and carefully generated negative examples.
Following this, we introduce our novel text encoder that is designed to leverage the unique structural properties of dialogue. 

\dtm{} is designed to output a scalar $\text{score}(c_i,r_i) = \mathcal{R} \in (0,1)$, which measures how appropriate a response $r_i$ is given a dialogue context $c_i$.
This task is analogous to measuring alignment in NLI, but instead of measuring entailment or contradiction, our notion of alignment aims to quantify the \emph{quality} of a dialogue response.  
As in NLI, we approach this task by defining encoders $f_e^{\theta}(c)$ and $f_e^{\theta}(r)$ to encode the context and response, a combination function $f_{comb}(.)$ to combine the representations, and a final classifier $f_t(.)$, which outputs the alignment score:
\begin{equation}
\label{eq:setup}
    \text{score}(c,r) = \sigma(f_t(f_{comb}(f_e^{\theta_1}(c),f_e^{\theta_2}(r))).
\end{equation}


The key idea behind an unreferenced dialogue metric is the use of Noise Contrastive Estimation (NCE) \cite{gutmann2010noise} for training. Specifically, we train the model to differentiate between a correct response ($\text{score}(c,r) \to 1$), and a negative response ($\text{score}(c,\hat{r}) \to 0$), where $\hat{r}$ represents a candidate false response for the given context $c$. The loss to minimize contains one positive example and a range of negative examples chosen from a sampling policy $P(\hat{r})$:
\begin{equation}
    \mathcal{L} = - \log(\text{score}(c,r)) - \mathbb{E}_{\hat{r} \sim P(\hat{r})} \log ( - \text{score}(c,\hat{r})).
\end{equation}
\noindent The sampling policy $P(\hat{r})$ consists of \textit{syntactic} and \textit{semantic} negative samples. 

\xhdr{Syntactic negative samples} We consider three variants of syntax level adversarial samples: \textit{word-order} (shuffling the ordering of the words of $r$), \textit{word-drop} (dropping $x$\% of words in $r$) and \textit{word-repeat} (randomly repeating words in $r$).

\xhdr{Semantic negative samples} We also consider three variants of negative samples that are syntactically well formed, but represent corruption in the semantic space.  First, we choose a response $r_j$ which is chosen at random from a different dialogue such that $r_j \ne r_i$ (\textit{random utterance}). Second, we use a pre-trained seq2seq model on the dataset, and pair random seq2seq generated response with $r_i$ (\textit{random seq2seq}). Third, to provide a bigger variation of semantically negative samples, for each $r_i$ we generate high-quality paraphrases $r_i^b$ using Back-Translation \cite{edunov2018backtranslation}. We pair random Back-Translations $r_j^b$ with $r_i$ as in the above setup (\textit{random back-translation}). We also provide the paired $r_i^b$ as positive example for the models to learn variation in semantic similarity.
We further discuss the effect of different sampling policies in Appendix \ref{sec:nce}.

\xhdr{Dialogue-structure aware encoder}
Traditional NLI approaches (e.g., \citet{conneauSupervisedLearningUniversal2017}) use the general setup of Equation \ref{eq:setup} to score context-response pairs. The encoder $f_e$ is typically a Bidirectional LSTM---or, more recently, a BERT-based model \cite{devlinBERTPretrainingDeep2018}, which uses a large pre-trained language model. $f_{comb}$ is defined as in \citet{conneauSupervisedLearningUniversal2017}:
\begin{equation}
\label{eq:comb}
    f_{comb}(u,v) = \text{concat}([u,v,u*v,u-v]).
\end{equation} 
However, the standard text encoders used in these traditional NLI approaches ignore the temporal structure of dialogues, which is critical in our setting where the context is composed of a sequence of distinct utterances, with natural and stereotypical transitions between them. (See Appendix \ref{sec:structure} for a qualitative analysis of these transitions).
Thus we propose a specialized text encoder for \dtm{}, which uses a BERT-based encoder $f_e^{\text{BERT}}$ but additionally models dialogue transitions using a recurrent neural network:



\begin{equation}
\label{eq:pool_eq}
\begin{split}
    \vh_{u_i} &= \mD_g f_e^{\text{BERT}}(u_i), \\
    \vh_{u_{i+1}}' &= f_R(\vh_{u_i}, \vh_{u_i}'), \\
    \vc_i &= \mW.\text{pool}_{\forall t \in \{u_1,\ldots,u_{n-1}\}}(\vh_{t}') \\
    \text{score}(c_i,r_i) &= \sigma (f_{t} ([\vh_{r_i}, \vc_{i}, \vh_{r_i}*\vc_i, \vh_{r_i}-\vc_{i}])),
\end{split}
\end{equation}

\noindent where $\vh_{u_i} \in \mathcal{R}^d$ is a downsampled BERT representation of the utterance $u_i$ (using a global learned mapping $\mD_g \in \mathcal{R}^{B\times d}$).  $\vh_{u_i}'$ is the hidden representation of $f_R$ for $u_i$, where $f_R$ is a Bidirectional LSTM.
The final representation of the dialogue context is learned by pooling the individual hidden states of the RNN using max-pool (Equation \ref{eq:pool_eq}). This context representation is mapped into the response vector space using weight $\mW$, to obtain $\vc_i$. We then learn the alignment score between the context $\vc_i$ and response $r_i$'s representation $\vh_{r_i}$ following Equation \ref{eq:setup}, by using the combination function $f_{comb}$ being the same as in Equation \ref{eq:comb}.


\section{Experiments}
To empirically evaluate our proposed unreferenced dialogue evaluation metric, we are interested in answering the following key research questions:

\begin{itemize}[itemsep=2pt, topsep=0pt]
    \item Q1: How robust is our proposed metric on different types of responses?
    \item Q2: How well does the self-supervised metric correlate with human judgements?
\end{itemize}

\xhdr{Datasets}  For training \dtm{}, we use PersonaChat \cite{zhang2018personalizing}, a large-scale open-domain chit-chat style dataset which is collected by human-human conversations over provided \textit{user persona}. We extract and process the dataset using ParlAI \cite{miller2017parlai} platform. We use the public train split for our training and validation, and the public validation split for testing. We use the human-human and human-model data collected by \citet{seeWhatMakesGood2019} for correlation analysis, where the models themselves are trained on PersonaChat.


\xhdr{Baselines} We use InferSent \cite{conneauSupervisedLearningUniversal2017} and unreferenced RUBER as  LSTM-based baselines. We also compare against BERT-NLI, which is the same as the InferSent model but with the LSTM encoder replaced with a pre-trained BERT encoder. 
Note that these baselines can be viewed as ablations of the \dtm{} framework using simplified text encoders, since we use the same NCE training loss to provide a fair comparison. 
Also, note that in practice, we use DistilBERT \cite{sanh2019distilbert} instead of BERT in both \dtm{} and the BERT-NLI baseline (and thus we refer to the BERT-NLI baseline as DistilBERT-NLI).\footnote{DistilBERT is the same BERT encoder with significantly reduced memory footprint and training time, which is trained by knowledge distillation \cite{buciluǎ2006model, hinton2015distilling} on the large pre-trained model of BERT.}.

\subsection{Evaluating \dtm{} on different types of responses}
\label{sec:ablations}

We first analyze the robustness of \dtm{} by comparing with the baselines, by using the same NCE training for \textit{all the models} for fairness. We evaluate the models on the \textit{difference} score, $\Delta = \text{score}(c,r_{\text{ground-truth}}) - \text{score}(c,r)$ (Table \ref{tab:ablation}). $\Delta$ provides an insight on the range of score function. An optimal metric would cover the full range of good and bad responses. We evaluate response $r$ in three settings: \textit{Semantic Positive}: responses that are semantically equivalent to the ground truth response; \textit{Semantic Negative}: responses that are semantically opposite to the ground truth response; and \textit{Syntactic Negative}: responses that have been adversarially modified in the lexical units.  Ideally, we would want $\Delta \to 1$ for semantic and syntactic negative responses, $\Delta \to 0$ for semantic positive responses. 

\begin{table}
\centering
\resizebox{\columnwidth}{!}{%
\begin{tabular}{|l|lllll|}
\hline
                                    &            & R & IS      & DNLI & M         \\ \hline
Semantic Positive $\downarrow$                   & BackTranslation  & 0.249 & 0.278          & \textbf{0.024} & 0.070          \\
                                    & Seq2Seq          & 0.342 & 0.362          & \textbf{0.174} & 0.308          \\ \hline
\multirow{2}{*}{Semantic Negative $\uparrow$}  & Random Utterance & 0.152 & 0.209          & 0.147          & \textbf{0.287} \\
                                    & Random Seq2Seq   & 0.402 & 0.435          & 0.344          & \textbf{0.585} \\ \hline
\multirow{3}{*}{Syntactic Negative $\uparrow$} & Word Drop        & 0.342 & \textbf{0.367} & 0.261          & 0.3            \\
                                    & Word Order       & 0.392 & 0.409          & 0.671          & \textbf{0.726} \\
                                    & Word Repeat      & 0.432 & 0.461          & 0.782          & \textbf{0.872} \\ \hline
\end{tabular}%
}
\caption{\small{Metric score evaluation ($\Delta = \text{score}(c,r_{\text{ground-truth}}) - \text{score}(c,r)$) between RUBER (R), InferSent (IS), DistilBERT-NLI (DNI) and \dtm{} (M) on PersonaChat dataset's public validation set. For Semantic Positive tests, lower $\Delta$ is better; for all Negative tests higher $\Delta$ is better.}}
\label{tab:ablation_short}
\end{table}




We observe that the \dtm{} scores perform robustly across all the setups. RUBER and InferSent baselines are weak, quite understandably so because they cannot leverage the large pre-trained language model data and thus is poor at generalization. DistilBERT-NLI baseline performs significantly better than InferSent and RUBER, while \dtm{} scores even better and more consistently overall. We provide a detailed ablation of various training scenarios as well as the absolute raw $\Delta$ scores in Appendix \ref{sec:nce}. We also observe both \dtm{} and DistilBERT-NLI to be more robust on zero-shot generalization to different datasets, the results of which are available in Appendix \ref{sec:generalization}.


\subsection{Correlation with human judgements}
\label{sec:human_judgements}

Metrics are evaluated on correlation with human judgements \cite{loweAutomaticTuringTest2017a, taoRUBERUnsupervisedMethod2017}, or by evaluating the responses of a generative model trained on the metric \cite{wietingBLEUTrainingNeural2019}, by human evaluation. However, this introduces a bias either during the questionnaire setup or during data post-processing in favor of the proposed metric. In this work, we refrain from collecting human annotations ourselves, but refer to the recent work by \citet{seeWhatMakesGood2019} on PersonaChat dataset. Thus, the evaluation of our metric is less subject to bias.


\citet{seeWhatMakesGood2019} conducted a large-scale human evaluation of 28 model configurations to study the effect of controllable attributes in dialogue generation. 
We use the publicly released model-human and human-human chat logs from \citet{seeWhatMakesGood2019} to generate the scores on our models, and correlate them with the associated human judgement on a Likert scale.
\citet{seeWhatMakesGood2019} propose to use a \textit{multi-step} evaluation methodology, where the human annotators rate the entire dialogue and not a context-response pair. On the other hand, our setup is essentially a \textit{single-step} evaluation method. To align our scores with the multi-turn evaluation, we average the individual turns to get an aggregate score for a given dialogue. 
\begin{figure}[!htb]
     \centering
    \includegraphics[width=0.5\textwidth]{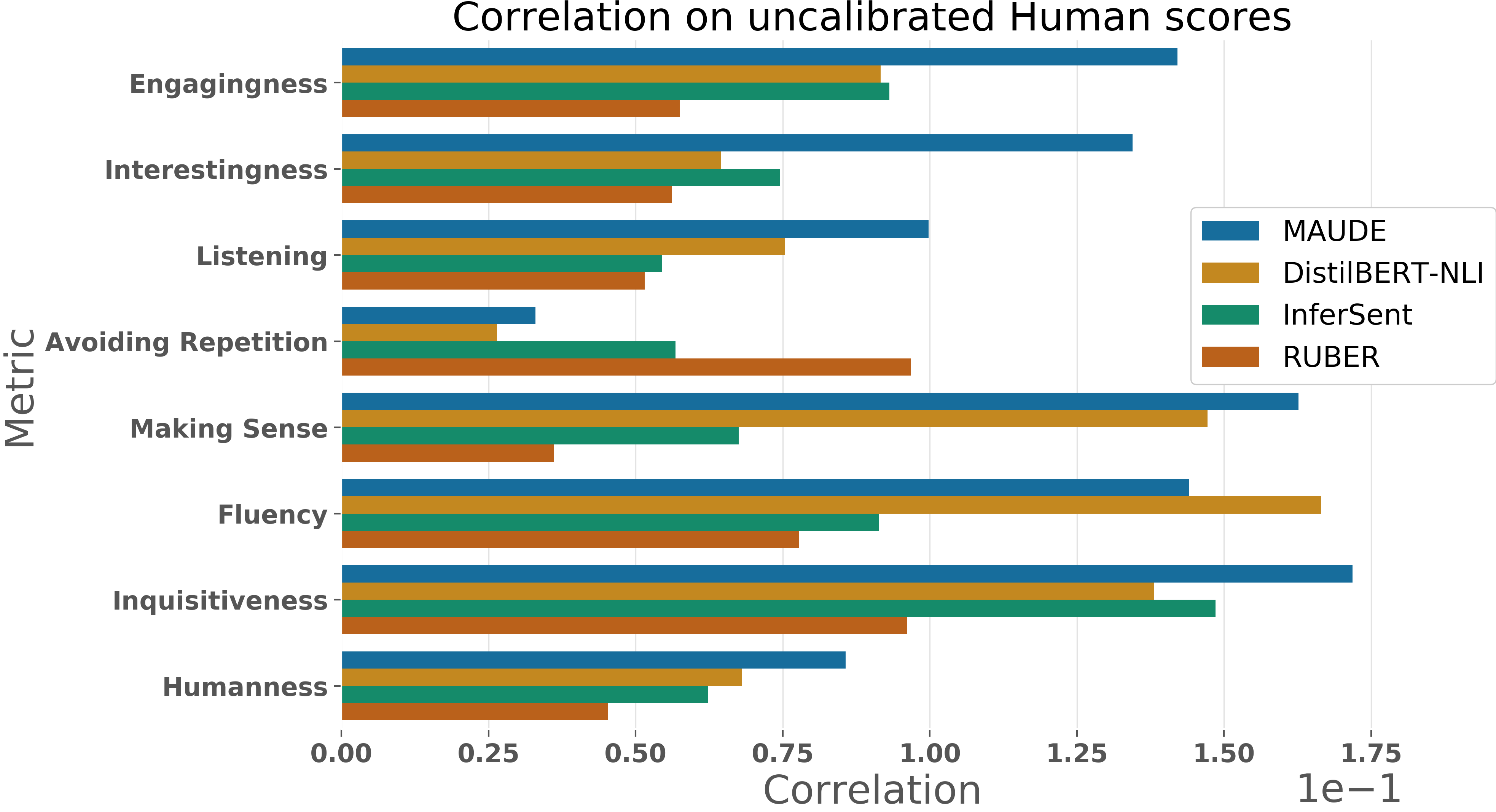}
    \caption{\small{Human correlation on un-calibrated scores collected on PersonaChat dataset \cite{zhang2018personalizing}, for \dtm{}, DistilBERT-NLI, InferSent and RUBER}}
    \label{fig:human_corr}
\end{figure}

\begin{table}[]
\centering
\resizebox{\columnwidth}{!}{%
\begin{tabular}{|l|l|l|l|l|}
\hline
                    & R          & IS & DNLI & M         \\ \cline{2-5} 
Fluency             & 0.322          & 0.246     & \textbf{0.443} & 0.37           \\
Engagingness        & 0.204          & 0.091     & 0.192          & \textbf{0.232} \\
Humanness           & 0.057          & -0.108    & \textbf{0.129} & 0.095          \\
Making Sense        & 0.0            & 0.005     & \textbf{0.256} & 0.208          \\
Inquisitiveness     & 0.583          & 0.589     & 0.598          & \textbf{0.728} \\
Interestingness     & \textbf{0.275} & 0.119     & 0.135          & 0.24           \\
Avoiding Repetition & \textbf{0.093} & -0.118    & -0.039         & -0.035         \\
Listening           & 0.061          & -0.086    & \textbf{0.124} & 0.112          \\ \hline
Mean                & 0.199          & 0.092     & 0.23           & \textbf{0.244} \\ \hline
\end{tabular}%
}
\caption{\small{Correlation with calibrated scores between RUBER (R), InferSent (IS), DistilBERT-NLI (DNI) and \dtm{} (M) when trained on PersonaChat dataset}}
\label{tab:human_corr_calib}
\vspace{-2mm}
\end{table}

We investigate the correlation between the scores and uncalibrated individual human scores from 100 crowdworkers (Fig. \ref{fig:human_corr}), as well as aggregated scores released by \citet{seeWhatMakesGood2019} which are adjusted for annotator variance by using Bayesian calibration \cite{kulikov2018importance} (Table \ref{tab:human_corr_calib}). In all cases, we report Spearman's correlation coefficients.

For uncalibrated human judgements, we observe \dtm{} having higher relative correlation in 6 out of 8 quality measures. Interestingly, in case of calibrated human judgements, DistilBERT proves to be better in half of the quality measures. \dtm{} achieves marginally better overall correlation for calibrated human judgements, due to significantly strong correlation on specifically two measures: Interestingness and Engagingness. These measures answers the questions \textit{``How interesting or boring did you find this conversation?"} and \textit{``How much did you enjoy talking to this user?"}. (Refer to Appendix B of \citet{seeWhatMakesGood2019} for the full list of questions). Overall, using large pre-trained language models provides significant boost in the human correlation scores.

\section{Conclusion}
In this work, we explore the feasibility of learning an automatic dialogue evaluation metric by leveraging pre-trained language models and the temporal structure of dialogue. We propose \dtm{}, which is an unreferenced dialogue evaluation metric that leverages sentence representations from large pre-trained language models, and is trained via Noise Contrastive Estimation. \dtm{} also learns a recurrent neural network to model the transition between the utterances in a dialogue, allowing it to correlate better with human annotations. This is a good indication that \dtm{} can be used to evaluate online dialogue conversations. Since it provides immediate continuous rewards and at the single-step level, \dtm{} can be also be used to optimize and train better dialogue generation models, which we want to pursue as future work.

\section*{Acknowledgements}
We would like to thank the ParlAI team (Margaret Li, Stephen Roller, Jack Urbanek, Emily Dinan, Kurt Shuster and Jason Weston) for technical help, feedback and encouragement throughout this project. We would like to thank Shagun Sodhani and Alborz Geramifard for helpful feedback on the manuscript. We would also like to thank William Falcon and the entire Pytorch Lightning community for making research code awesome. We are grateful to Facebook AI Research (FAIR) for providing extensive compute / GPU resources and support regarding the project. This research, with respect to Quebec Artificial Intelligence Institute (Mila) and McGill University, was supported by the Canada CIFAR Chairs in AI program. 

\bibliography{acl2020}
\bibliographystyle{acl_natbib}

\clearpage
\appendix
\section{Temporal Structure}
\label{sec:structure}

We hypothesize that a \emph{good} encoding function can capture the structure that exists in dialogue. Often this translates to capturing the semantics, coherency in dialogue which are some of the key attributes of a conversation. Formally, we propose using a function $f_t^{D_i}$ which maps one utterance to the next.

\begin{figure*}[h]
     \centering
    \subfloat{{\includegraphics[width=0.24\textwidth]{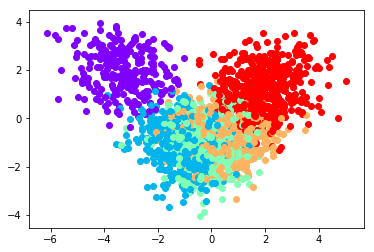}}}%
    \qquad
    \hspace{-20pt}
    \subfloat{{\includegraphics[width=0.24\textwidth]{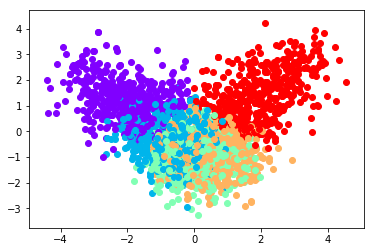}}}%
    \qquad
    \hspace{-25pt}
    \subfloat{{\includegraphics[width=0.24\textwidth]{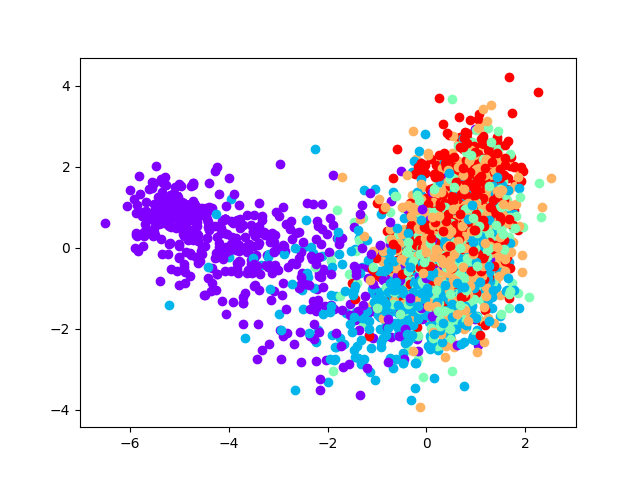}}}%
    \qquad
    \hspace{-30pt}
    \subfloat{{\includegraphics[width=0.24\textwidth]{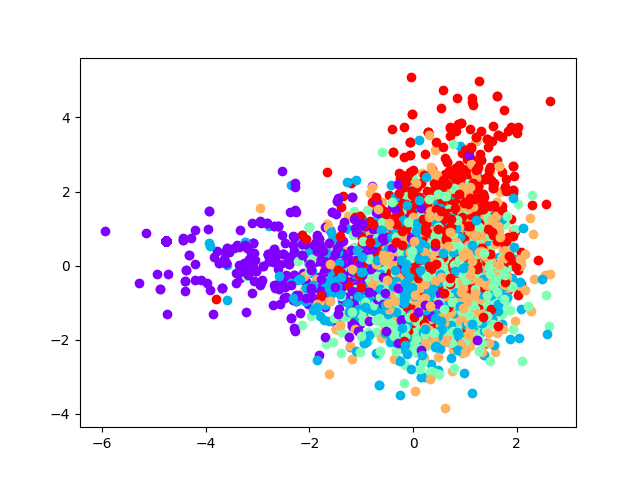}}}%
    \caption{From left to right, LDA downsampled representation of BERT on Frames (Goal oriented), MultiWOZ (Goal oriented), PersonaChat (chit-chat) and DailyDialog (chit-chat)}
    \vspace{-10pt}
    \label{fig:dialog_structure}
\end{figure*}

\begin{equation}
  \vh_{u_{i+1}} = f_t^{D_i}(\vh_{u_i})
\end{equation}

To define a \textit{good} encoding function, we turn to pre-trained language models. These models are typically trained on large corpus and achieve state-of-the-art results on a range of language understanding tasks \cite{ott2018scaling}.
To validate our hypothesis, we use a pre-trained (and fine-tuned) BERT \cite{devlinBERTPretrainingDeep2018} as $f_e$. We compute $h_{u_i} = f_e(u_i) \forall u_i \in D$, and learn a linear classifier to predict an approximate position of the $u_i \in D_i$. The task has details in its design, in the case of goal-oriented dialogues the vocabulary differs in different parts of the conversation and in chitchat dialogues it cannot be said. 
To experiment, we choose PersonaChat \cite{zhang2018personalizing} and DailyDialog \cite{li2017dailydialog} to be nominal of chit-chat style data, and Frames \cite{asri2017frames} and MultiWOZ \cite{budzianowski2018multiwoz} for goal-oriented data. 

We encode every consecutive pairs of the utterances with a \% score, \emph{t}, that denotes its occurrence after the completion of \emph{t}\% of dialogue.
 \begin{equation}
     t_{u_p} = \frac{\text{index}_{u_p}+1}{k}
 \end{equation}
 where $\text{index}_{u_p}$ denote the average of the indices in the pair of the utterances and \emph{k} denote the total number of utterances in dialogue.

 Now, we pre-define the number of bins \emph{B}. We split the range 0-100 into \emph{B} non-overlapping sets(every set will have min and max denoted by $s^i_{min}$ and $s^i_{max}$ respectively). We parse every dialogue in the dataset, and place the encoding of every utterance pair in the corresponding bin.

 \begin{equation}
     bin_{u_p} = \{i \mid t_{u_p} > s^i_{min} \& s^i_{max} > t_{u_p}\}
 \end{equation}

We then use Linear Discriminant Analysis (LDA) to predict the bin of each utterance $u_i$ in the dialogue after converting the high dimensional embedding into 2 dimensions. LDA provides the best possible class conditioned representation of data. This gives us a downsampled representation of each utterance $u_i$ which we plot as shown in Figure \ref{fig:dialog_structure}.  The reduction on BERT encoding to 2-dimensions shows that BERT is useful in nudging the encoded utterances towards useful structures. We see well defined clusters in goal-oriented but not-so-well-defined clusters in open domain dialogues. This is reasonable to expect and intuitive. 

\section{Generalization on unseen dialog datasets}
\label{sec:generalization}
\begin{table*}[]
\centering
\resizebox{\textwidth}{!}{%
\begin{tabular}{|l|ccl|ll|ll|}
\hline
Datasets & \multicolumn{1}{l|}{} & \multicolumn{2}{l|}{DailyDialog} & \multicolumn{2}{l|}{Frames} & \multicolumn{2}{l|}{MultiWOZ} \\ \hline
Model & \multicolumn{1}{c}{Eval Mode} & \multicolumn{1}{c}{Score} & $\Delta$ & \multicolumn{1}{l}{Score} & $\Delta$ & \multicolumn{1}{l}{Score} & $\Delta$ \\ \hline
\multirow{2}{*}{RUBER} 
& $+$ & 0.173 $\pm{0.168}$ &  & 0.211 $\pm{0.172}$ &  & 0.253 $\pm{0.177}$ &  \\
 & $-$ & 0.063 $\pm{0.092}$ & 0.11 & 0.102 $\pm{0.114}$ & 0.109 & 0.121 $\pm{0.123}$ & 0.123 \\ \hline
\multirow{2}{*}{InferSent} 
& $+$ & 0.163 $\pm{0.184}$ &  & 0.215 $\pm{0.186}$ &  & 0.277 $\pm{0.200}$ &  \\
 & $-$ & 0.050 $\pm{0.085}$ & 0.113 & 0.109 $\pm{0.128}$ & 0.106 & 0.127 $\pm{0.133}$ & 0.15 \\ \hline
\multirow{2}{*}{DistilBERT NLI} 
& $+$ & 0.885 $\pm{0.166}$ &  & 0.744 $\pm{0.203}$ &  & 0.840 $\pm{0.189}$ &  \\
 & $-$ & 0.575 $\pm{0.316}$ & 0.31 & 0.538 $\pm{0.330}$ & 0.206 & 0.566 $\pm{0.333}$ & 0.274 \\ \hline
\multirow{2}{*}{\dtm{}}
& $+$ & 0.782 $\pm{0.248}$ &  & 0.661 $\pm{0.293}$ &  & 0.758 $\pm{0.265}$ & \\ 
& $-$ & 0.431 $\pm{0.300}$ & \textbf{0.351} & 0.454 $\pm{0.358}$ & \textbf{0.207} & 0.483 $\pm{0.345}$ & \textbf{0.275} \\ \hline
\end{tabular}%
}
\caption{Zero-shot generalization results on DailyDialog, Frames and MultiWOZ dataset for the baselines and \dtm{}. $+$ denotes semantic positive responses, and $-$ denotes semantic negative responses.}
\label{tab:zero_shot}
\end{table*}

In order for a dialogue evaluation metric to be useful, one has to evaluate how it generalizes to unseen data. We performed the evaluation using our trained models on PersonaChat dataset, and then evaluated them \textit{zero-shot} on two goal-oriented datasets, Frames \cite{asri2017frames} and MultiWoz \cite{budzianowski2018multiwoz}, and one chit-chat style dataset: Daily Dialog \cite{li2017dailydialog} (Table \ref{tab:zero_shot}). We find BERT-based models are significantly better at generalization than InferSent or RUBER, with \dtm{} marginally better than DistilBERT-NLI baseline. \dtm{} has the biggest impact on generalization to DailyDialog dataset, which suggests that it captures the commonalities of chit-chat style dialogue from PersonaChat. Surprisingly, generalization gets significantly better of BERT-based models on goal-oriented datasets as well. This suggests that irrespective of the nature of dialogue, pre-training helps because it contains the information common to English language lexical items.

\section{Noise Contrastive Estimation training ablations}
\label{sec:nce}

\begin{table*}[]
\centering
\resizebox{\textwidth}{!}{%
\begin{tabular}{|l|l|ll|ll|ll|ll|}
\hline
PersonaChat Dataset & Model & \multicolumn{2}{c|}{RUBER} & \multicolumn{2}{c|}{InferSent} & \multicolumn{2}{c|}{DistilBERT NLI} & \multicolumn{2}{c|}{\dtm{}} \\ \cline{2-10} 
 & Training Modes & \multicolumn{2}{c|}{Only Semantics} & \multicolumn{2}{c|}{Only Semantics} & \multicolumn{2}{c|}{Only Semantics} & \multicolumn{2}{c|}{Only Semantics} \\ \cline{2-10} 
 & Evaluation Modes & \multicolumn{1}{l|}{Score} & $\Delta$ & \multicolumn{1}{l|}{Score} & $\Delta$ & \multicolumn{1}{l|}{Score} & $\Delta$ & \multicolumn{1}{l|}{Score} & $\Delta$ \\ \hline
\multirow{3}{*}{Semantic Positive} & Gold Truth Response & 0.443\tiny$\pm{0.197}$ & 0 & 0.466\tiny$\pm{0.215}$ & 0 & 0.746\tiny$\pm{0.236}$ & 0 & \textbf{0.789}\tiny$\pm{0.244}$ & 0 \\
 & BackTranslation & 0.296\tiny$\pm{0.198}$ & 0.147 & 0.273\tiny$\pm{0.195}$ & 0.192 & \textbf{0.766}\tiny$\pm{0.235}$ & -0.02 & 0.723\tiny$\pm{0.277}$ & \boldblue{0.066} \\
 & Seq2Seq & 0.082\tiny$\pm{0.163}$ & 0.361 & 0.10\tiny$\pm{0.184}$ & 0.367 & \textbf{0.46}\tiny$\pm{0.357}$ & \boldblue{0.286} & 0.428\tiny$\pm{0.390}$ & 0.361 \\ \hline
\multirow{2}{*}{Semantic Negative} & Random Utterance & 0.299\tiny$\pm{0.203}$ & 0.144 & \textbf{0.287}\tiny$\pm{0.208}$ & 0.178 & 0.489\tiny$\pm{0.306}$ & 0.257 & 0.388\tiny$\pm{0.335}$ & \boldblue{0.40} \\
 & Random Seq2Seq & \textbf{0.028}\tiny$\pm{0.077}$ & 0.415 & 0.036\tiny$\pm{0.082}$ & 0.429 & 0.237\tiny$\pm{0.283}$ & 0.529 & 0.16\tiny$\pm{0.26}$ & \boldblue{0.629} \\ \hline 
\multirow{3}{*}{Syntactic Negative} & Word Drop & 0.334\tiny$\pm{0.206}$ & 0.109 & \textbf{0.308}\tiny$\pm{0.217}$ & \boldblue{0.158} & 0.802\tiny$\pm{0.224}$ & -0.056 & 0.73\tiny$\pm{0.29}$ & 0.059 \\
 & Word Order & \textbf{0.472}\tiny$\pm{0.169}$ & -0.029 & 0.482\tiny$\pm{0.19}$ & -0.016 & 0.685\tiny$\pm{0.284}$ & 0.061 & 0.58\tiny$\pm{0.35}$ & \boldblue{0.209} \\
 & Word Repeat & 0.255\tiny$\pm{0.24}$ & 0.188 & \textbf{0.153}\tiny$\pm{0.198}$ & 0.312 & 0.657\tiny$\pm{0.331}$ & 0.089 & 0.44\tiny$\pm{0.39}$ & \boldblue{0.349} \\ \hline
\end{tabular}%
}
\caption{Metric score evaluation between InferSent, DistilBERT-NLI and \dtm{} on PersonaChat dataset, trained on $P(\hat{r})$ = Semantics. Bold scores represent the best individual scores, and bold with blue represents the best difference with the true response.}
\label{tab:ablation_semantics}
\end{table*}

\begin{table*}[]
\centering
\resizebox{\textwidth}{!}{%
\begin{tabular}{|l|l|ll|ll|ll|ll|}
\hline
PersonaChat Dataset & Model & \multicolumn{2}{c|}{RUBER} & \multicolumn{2}{c|}{InferSent} & \multicolumn{2}{c|}{DistilBERT NLI} & \multicolumn{2}{c|}{\dtm{}} \\ \cline{2-10} 
 & Training Modes & \multicolumn{2}{c|}{Only Syntax} & \multicolumn{2}{c|}{Only Syntax} & \multicolumn{2}{c|}{Only Syntax} & \multicolumn{2}{c|}{Only Syntax} \\ \cline{2-10} 
 & Evaluation Modes & \multicolumn{1}{l|}{Score} & $\Delta$ & \multicolumn{1}{l|}{Score} & $\Delta$ & \multicolumn{1}{l|}{Score} & $\Delta$ & \multicolumn{1}{l|}{Score} & $\Delta$ \\ \hline
\multirow{3}{*}{Semantic Positive} & Gold Truth Response & 0.891\tiny$\pm{0.225}$ & 0 & 0.893\tiny$\pm{0.231}$ & 0 & 0.986\tiny$\pm{0.088}$ & 0 & \textbf{0.99}\tiny$\pm{0.07}$ & 0 \\
 & BackTranslation & 0.687\tiny$\pm{0.363}$ & 0.204 & 0.672\tiny$\pm{0.387}$ & 0.221 & 0.877\tiny$\pm{0.268}$ & 0.109 & \textbf{0.91}\tiny$\pm{0.23}$ & \boldblue{0.08} \\
 & Seq2Seq & 0.929\tiny$\pm{0.187}$ & -0.038 & 0.949\tiny$\pm{0.146}$ & -0.055 & 0.996\tiny$\pm{0.048}$ & -0.01 & \textbf{0.99}\tiny$\pm{0.05}$ & \boldblue{0.00} \\ \hline
\multirow{2}{*}{Semantic Negative} & Random Utterance & 0.869\tiny$\pm{0.248}$ & 0.022 & \textbf{0.835}\tiny$\pm{0.294}$ & \boldblue{0.058} & 0.977\tiny$\pm{0.116}$ & 0.009 & 0.97\tiny$\pm{0.13}$ & 0.02 \\
 & Random Seq2Seq & 0.915\tiny$\pm{0.196}$ & -0.024 & \textbf{0.904}\tiny$\pm{0.206}$ & -0.011 & 0.994\tiny$\pm{0.057}$ & -0.008 & 0.99\tiny$\pm{0.08}$ & \boldblue{0} \\ \hline
\multirow{3}{*}{Syntactic Negative} & Word Drop & 0.119\tiny$\pm{0.255}$ & 0.772 & \textbf{0.105}\tiny$\pm{0.243}$ & \boldblue{0.788} & 0.373\tiny$\pm{0.414}$ & 0.613 & 0.41\tiny$\pm{0.44}$ & 0.584 \\
 & Word Order & 0.021\tiny$\pm{0.101}$ & 0.87 & \textbf{0.015}\tiny$\pm{0.0915}$ & 0.878 & 0.064\tiny$\pm{0.194}$ & 0.922 & 0.07\tiny$\pm{0.21}$ & \boldblue{0.928} \\
 & Word Repeat & \textbf{0.001}\tiny$\pm{0.007}$ & 0.89 & \textbf{0.001}\tiny$\pm{0.020}$ & 0.893 & 0.006\tiny$\pm{0.057}$ & 0.980 & 0.01\tiny$\pm{0.06}$ & \boldblue{0.981} \\ \hline
\end{tabular}%
}
\caption{Metric score evaluation between InferSent, DistilBERT-NLI and \dtm{} on PersonaChat dataset, trained on $P(\hat{r})$ = Syntax. Bold scores represent the best individual scores, and bold with blue represents the best difference with the true response.}
\label{tab:ablation_syntax}
\end{table*}

\begin{table*}[ht]
\centering
\resizebox{\textwidth}{!}{%
\begin{tabular}{|l|l|ll|ll|ll|ll|}
\hline
PersonaChat Dataset & Model & \multicolumn{2}{c}{RUBER} & \multicolumn{2}{c|}{InferSent} & \multicolumn{2}{c|}{DistilBERT NLI} & \multicolumn{2}{c|}{\dtm{}} \\ \cline{2-10} 
 & Training Modes & \multicolumn{2}{c}{All} & \multicolumn{2}{c|}{All} & \multicolumn{2}{c|}{All} & \multicolumn{2}{c|}{All} \\ \cline{2-10} 
 & Evaluation Modes & \multicolumn{1}{l|}{Score} & $\Delta$ & \multicolumn{1}{l|}{Score} & $\Delta$ & \multicolumn{1}{l|}{Score} & $\Delta$ & \multicolumn{1}{l|}{Score} & $\Delta$ \\ \hline
\multirow{3}{*}{Semantic Positive} & Gold Truth Response & 0.432\tiny$\pm{0.213}$ & 0 & 0.462\tiny$\pm{0.254}$ & 0 & 0.824\tiny$\pm{0.154}$ & 0 & \textbf{0.909}\tiny$\pm{0.152}$ & 0 \\
 & BackTranslation & 0.183\tiny$\pm{0.198}$ & 0.249 & 0.184\tiny$\pm{0.218}$ & 0.278 & 0.8\tiny$\pm{0.19}$ & \boldblue{0.024} & \textbf{0.838}\tiny$\pm{0.227}$ & 0.070 \\
 & Seq2Seq & 0.09\tiny$\pm{0.17}$ & 0.342 & 0.10\tiny$\pm{0.184}$ & 0.362 & \textbf{0.65}\tiny$\pm{0.287}$ & \boldblue{0.174} & 0.6008\tiny$\pm{0.38}$ & 0.308 \\ \hline
\multirow{2}{*}{Semantic Negative} & Random Utterance & 0.28\tiny$\pm{0.21}$ & 0.152 & \textbf{0.252}\tiny$\pm{0.236}$ & 0.209 & 0.677\tiny$\pm{0.255}$ & 0.147 & 0.621\tiny$\pm{0.344}$ & \boldblue{0.287} \\
 & Random Seq2Seq & 0.03\tiny$\pm{0.09}$ & 0.402 & \textbf{0.026}\tiny$\pm{0.079}$ & 0.435 & 0.48\tiny$\pm{0.313}$ & 0.344 & 0.323\tiny$\pm{0.355}$ & \boldblue{0.585} \\ \hline 
\multirow{3}{*}{Syntactic Negative} & Word Drop & \textbf{0.09}\tiny$\pm{0.16}$ & 0.342 & 0.094\tiny$\pm{0.17}$ & \boldblue{0.367} & 0.563\tiny$\pm{0.377}$ & 0.261 & 0.609\tiny$\pm{0.401}$ & 0.3 \\
 & Word Order & \textbf{0.04}\tiny$\pm{0.10}$ & 0.392 & 0.052\tiny$\pm{0.112}$ & 0.409 & 0.153\tiny$\pm{0.29}$ & 0.671 & 0.182\tiny$\pm{0.327}$ & \boldblue{0.726} \\
 & Word Repeat & \textbf{0.00}\tiny$\pm{0.01}$ & 0.432 & 0.001\tiny$\pm{0.010}$ & 0.461 & 0.041\tiny$\pm{0.153}$ & 0.782 & 0.036\tiny$\pm{0.151}$ & \boldblue{0.872} \\ \hline
\end{tabular}
}
\caption{Metric score evaluation between InferSent, DistilBERT-NLI and \dtm{} on PersonaChat dataset, trained on $P(\hat{r})$ = Syntax + Semantics. Bold scores represent the best individual scores, and bold with blue represents the best difference with the true response.}
\label{tab:ablation}
\end{table*}

The choice of negative samples (Section \ref{sec:model}) for Noise Contrastive Estimation can have a large impact on the test-time scores of the metrics. In this section, we show the effect when we train only using syntactic negative samples (Table \ref{tab:ablation_semantics}) and only semantic negative samples (Table \ref{tab:ablation_syntax}). For comparison, we show the full results when trained using both of the sampling scheme in Table \ref{tab:ablation}.  We find overall training only using either syntactic or semantic negative samples achieve less $\Delta$ than training using both of the schemes. All models achieve high scores on the semantic positive samples when only trained with syntactical adversaries. However, training only with syntactical negative samples results in adverse effect on detecting semantic negative items.


\section{Qualitative Evaluation}
\label{sec:qualitative evaluation}

\begin{figure*}[!htb]
     \centering
    \includegraphics[width=\textwidth]{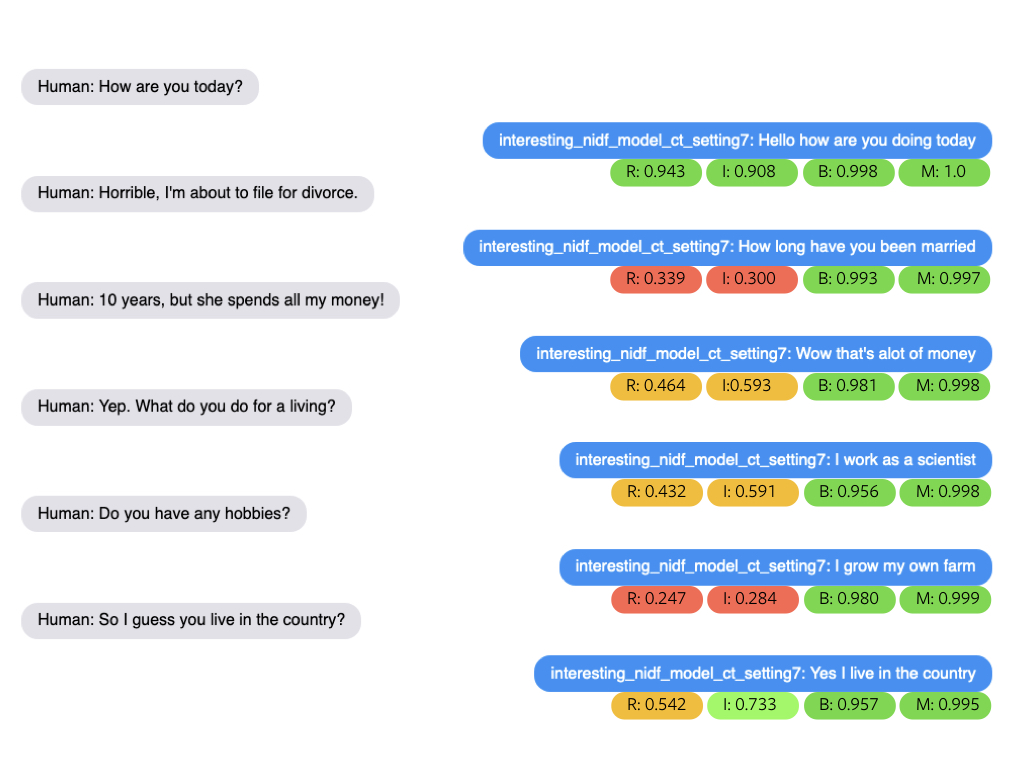}
    \vspace{-10pt}
    \caption{An example of dialogue conversation between human and a strong model, where \dtm{} (M) score correlates positively with human annotations. Raw Likert scores for the entire dialogue are: Engagingness : 3, Interestingness : 3, Inquisitiveness : 2, Listening : 3, Avoiding Repetition : 3, Fluency : 4, Making Sense : 4, Humanness : 3, Persona retrieval : 1. Baselines are RUBER (R), InferSent (I) and BERT-NLI  (B). }
    \label{fig:good_dial_high}
\end{figure*}

\begin{figure*}[!htb]
     \centering
    \includegraphics[width=\textwidth]{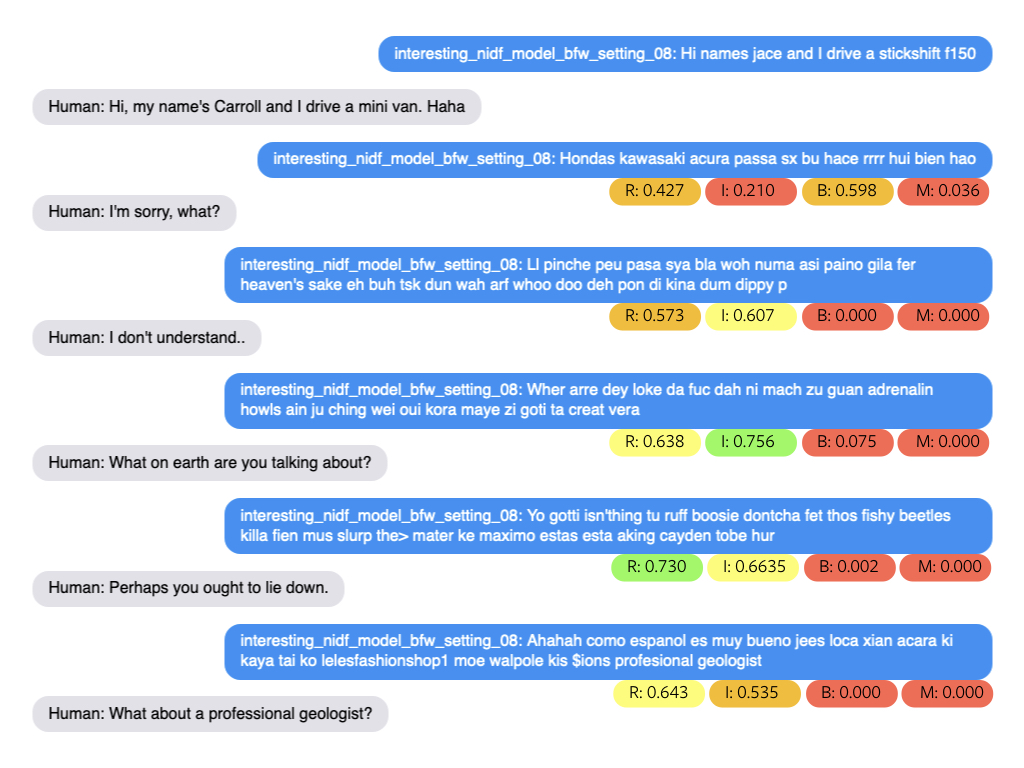}
    \vspace{-10pt}
    \caption{An example of dialogue conversation between human and a weak model, where \dtm{} (M) score correlates positively with human annotations. Raw Likert scores for the entire dialogue are: Engagingness : 1, Interestingness : 4, Inquisitiveness : 1, Listening : 1, Avoiding Repetition : 3, Fluency : 1, Making Sense : 2, Humanness : 1, Persona retrieval : 1. In our setup we only score responses \textit{only} following a human response. Baselines are RUBER (R), InferSent (I) and BERT-NLI  (B).}
    \label{fig:good_dial_low}
\end{figure*}

\begin{figure*}[!htb]
     \centering
    \includegraphics[width=\textwidth]{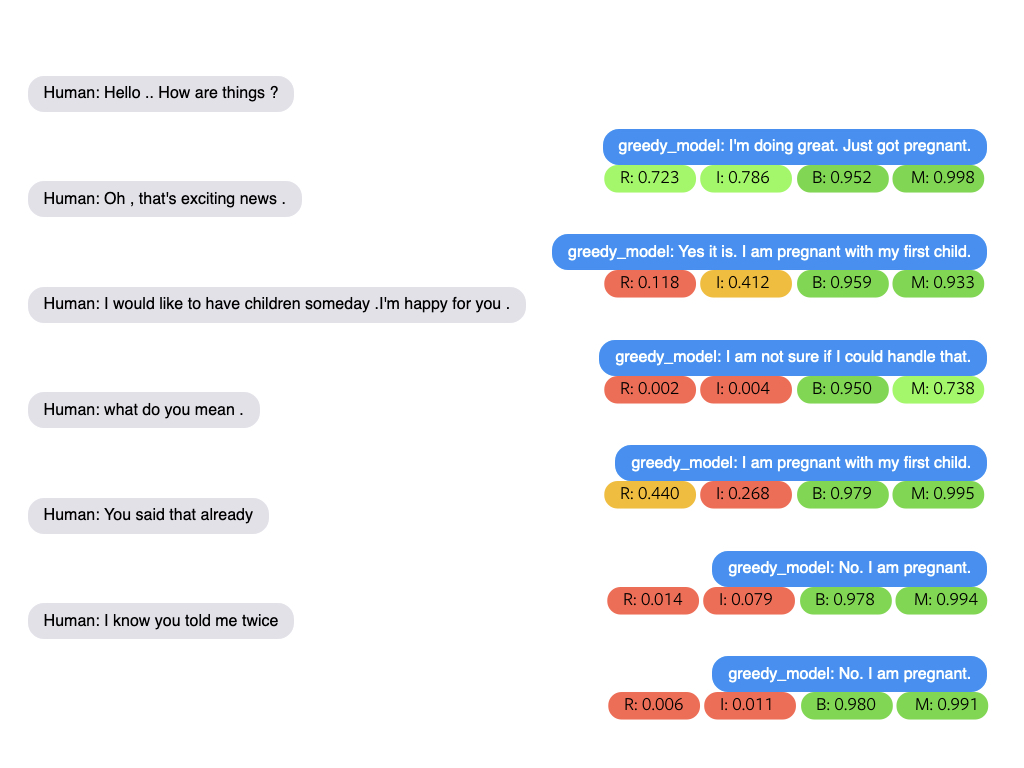}
    \vspace{-10pt}
    \caption{An example of dialogue conversation between human and a model, where \dtm{} (M) score correlates negatively with human annotations. Raw Likert scores for the entire dialogue are: Engagingness : 1, Interestingness : 1, Inquisitiveness : 2, Listening : 2, Avoiding Repetition : 2, Fluency : 3, Making Sense : 4, Humanness : 2, Persona retrieval : 1. Baselines are RUBER (R), InferSent (I) and BERT-NLI  (B).}
    \label{fig:bad_dial}
\end{figure*}

We investigate qualitatively how the scores of different models are on the online evaluation setup on \citet{seeWhatMakesGood2019}'c collected data. In Figure \ref{fig:good_dial_high}, we show a sample conversation where a human evaluator is pitched against a strong model. Here, \dtm{} scores correlate strongly with raw likert scores on different metrics. We observe that RUBER and InferSent baselines overall correlate negatively with the response. In Figure \ref{fig:good_dial_low}, we show another sample where a human evaluator is pitched against a weak model, which exhibits degenerate responses. We see both \dtm{} and DistilBERT-NLI correlate strongly with human annotation and provides a very low score, compared to RUBER or InferSent.

Since we essentially cherry-picked good results, its only fair to show a similarly cherry-picked negative example of \dtm{}. We sampled from responses where \dtm{} scores are negatively correlated with human annotations on Inquisitiveness metric (5\% of cases), and we show one of those responses in Figure \ref{fig:bad_dial}. We notice how both DistilBERT-NLI and \dtm{} fails to recognize the duplication of utterances which leads to a low overall score. This suggests there still exists room for improvement in developing \dtm{}, possibly by training the model to detect degeneracy in the context.

\section{Hyperparameters and Training Details}
\label{sec:hyperparams}

We performed rigorous hyperparameter search to tune our model \dtm{}. We train \dtm{} with downsampling, as we observe poor results when we run the recurrent network on top of 768 dimensions. Specifically, we downsample to 300 dimensions, which is the same used by our baselines RUBER and InferSent in their respective encoder representations. We also tested with the choice of either learning a PCA to downsample the BERT representations vs learning the mapping $\mD_g$ (Equation \ref{eq:pool_eq}), and found the latter producing better results. We keep the final decoder same for all models, which is a two layer MLP with hidden layer of size 200 dimensions and dropout 0.2. For BERT-based models (DistilBERT-NLI and \dtm{}), we use HuggingFace Transformers \cite{Wolf2019HuggingFacesTS} to first fine-tune the training dataset on language model objective. We tested with training on frozen fine-tuned representations in our initial experiments, but fine-tuning end-to-end lead to better ablation scores. For all models we train using Adam optimizer with 0.0001 as the learning rate, early stopping till validation loss doesn't improve. For the sake of easy reproducibility, we use Pytorch Lightning \cite{Falcon2019} framework. We used 8 Nvidia-TitanX GPUs on a DGX Server Workstation to train faster using Pytorch Distributed Data Parallel (DDP).

\end{document}